\documentclass[journal]{IEEEtran}
\usepackage{CJKutf8}
\usepackage{graphicx}
\usepackage{enumerate}
\usepackage{url}
\usepackage{diagbox}

\begin{document}
\begin{CJK*}{UTF8}{song}

\title{A Survey of Hybrid Human-Artificial Intelligence for Social Computing}

\author{Wenxi Wang,
        Huansheng Ning,~\IEEEmembership{Senior Member,~IEEE,}
        Feifei Shi, Sahraoui Dhelim, Weishan zhang
        \\and Liming Chen% <-this % stops a space}
\thanks{W. Wang is with the School of Computer and Communication Engineering, University of Science and Technology Beijing, Beijing 100083, China.}% <-this % stops a space

\thanks{H. Ning is with the School of Computer and Communication Engineering, University of Science and Technology Beijing, Beijing 100083, China, and also with the Beijing Engineering Research Center for Cyberspace Data Analysis and Applications, Beijing 100083, China. }% <-this % stops a space

\thanks{F. Shi is with the School of Computer and Communication Engineering, University of Science and Technology Beijing, Beijing 100083, China.}
\thanks{S. Dhelim is with the School of Computer and Communication Engineering, University of Science and Technology Beijing, Beijing 100083, China.}
\thanks{W. Zhang is with the Department of Software Engineering, China University of Petroleum, Qingdao, 266580, China.}
\thanks{L. Chen is with the School of Computing at Ulster University, UK.}}

\maketitle
\begin{abstract}
Along with the development of modern computing technology and social sciences, both theoretical research and practical applications of social computing have been continuously extended. In particular with the boom of artificial intelligence (AI), social computing is significantly influenced by AI. However, the conventional technologies of AI have drawbacks in dealing with more complicated and dynamic problems. Such deficiency can be rectified by hybrid human-artificial intelligence (H-AI) which integrates both human intelligence and AI into one unity, forming a new enhanced intelligence. H-AI in dealing with social problems shows the advantages that AI can't surpass. This paper firstly introduces the concepts of H-AI. AI is the intelligence in the transition stage of H-AI, so the latest research progresses of AI in social computing are reviewed. Secondly, it summarizes typical challenges faced by AI in social computing, and makes it possible to introduce H-AI to solve these challenges. Finally, the paper proposes a holistic framework of social computing combining with H-AI, which consists of four layers: object layer, base layer, analysis layer, and application layer. It represents H-AI has significant advantages over AI in solving social problems.
\end{abstract}


\IEEEpeerreviewmaketitle

\section{Introduction}

%ÏÖÔÚ£¬Éç»áýÌåµÄ¿ìËÙ·¢Õ¹Ê¹µÃ¸ü¶àµÄÈË¿ªÊ¼»îÔ¾ÔÚÍøÂçÉÏ£¬²úÉúÁËÁ˾޴óµÄÊý¾ÝÁ¿£¬ÕâЩÊý¾ÝÊÇÑо¿ÈËÀàÉç»á¡¢·¢Õ¹È˹¤ÖÇÄܼ¼ÊõµÄÖØÒª×ÊÔ´¡£´Ó½»ÓÑ¡¢Òûʳ¡¢³öÐе½ÓéÀÖ¡¢½ÌÓý¡¢Í¶×ʵȸ÷¸öÉú»îÁìÓò¶¼ÒѾ­¿ªÕ¹ÁËÏßÉÏÒµÎñ£¬²»½ö½öÔì³ÉÁËÐÅÏ¢±¬Õ¨£¬ÐÅÏ¢µÄά¶ÈÒ²Öð½¥±äµÃ·á¸»£¬ÀûÓüÆËã·½·¨´¦ÀíÉ罻ýÌåÊý¾Ý£¬¿ÉÒÔ±ÈÒÔÍù¸ü¼ÓÓÐЧµÄ½øÐÐÉç»áѧ·½ÃæµÄÑо¿£¬Éç»á¼ÆËãÒ²Òò´ËÐËÆð¡£Éç»á¼ÆËãÁìÓòÊÇÒ»¸öÓÐ׿«´ó·¢Õ¹Ç±Á¦µÄÐÂÐËÁìÓò£¬µ±½ñÈËÀàÉç»á±»ÐÅÏ¢¼¼ÊõÒÔÒ»ÖÖ¸ü¼Ó¿ìËÙ¸ßЧµÄ·½Ê½×éÖ¯ÔÚÒ»Æ𣬾޴óµÄÉç»áÊý¾Ý´æ´¢ÔÚ»¥ÁªÍøÖУ¬µÈ´ý×ÅÎÒÃÇÈ¥ÍÚ¾òÆäÖеı¦²Ø£¬ÕâÒ²´ÙʹÁËÈË»úÈÚºÏÖÇÄÜÔÚÉç»á¼ÆËãÖеÄÓ¦ÓÃÓë·¢Õ¹¡£

\IEEEPARstart{N}owadays, with the rapid development of social media, numerous people begin to be active on the network, which produces a huge amount of data. These data are important resources for the research of human society and the development of artificial intelligence technology. From making friends, diet, travel to entertainment, education, investment and other fields of life, online business has been carried out, which not only causes the information explosion, but also gradually enriches the dimensions of information. It is more effective to carry out sociological research relying on computational methods to deal with social media data. Social computing has also sprung up. The field of social computing is an emerging field with great development potential. Human society is organized by information technology in a more rapid and efficient way. Massive social data is stored in the Internet, waiting for us to excavate the treasure, which also promotes the application and development of hybrid human-artificial intelligent (H-AI) in social computing.

\par In 1994, the concept of social computing was first proposed by Schuler \cite{ref2}. He thought, "Social computing can be of computing application, with software as the medium or focus of social relationships." Charron et al. defines social computing as a technology that affects an individual or a community rather than an institution's social architecture. Wang et al. \cite{ref3} gives definitions of it from general and narrow levels. In a broad sense, social computing refers to computational theories and methods for social science. While in a narrow sense, it represents social activities, processes, structures, organizations and the calculation of their roles and effects. Dryer et al.\cite{ref4} describes social computing from scientific and technical aspects, that is, "social computing is generated by the social and interactive behaviors of human using computing technology." However, there is no clear definition of social computing as a cross-integration of social sciences and computational sciences.

\par The technology applied in social computing is a part of AI, which has experienced at least two waves in its development. The first wave appeared between 1956 and 1974, when AI took a knowledge-based approach, and machines with intelligence were produced. In the second wave, which appeared in the 1980s to the end of the 20th century, AI spawned a multi-layer neural network and BP back-propagation algorithm, and developed a highly intelligent machine AlphaGo that can play chess with humans. With the popularity of the Internet, the ubiquity of senors, the emergence of big data, the development of e-commerce, and the information society, data and knowledge are intertwined and interacted between human society, physical space and cyberspace. It is the third wave. AI tries to realize the cooperation between human and machine through big data and gradually upgraded algorithm. The integration of human-artificial intelligence (H-AI) will be the next breakthrough point in the future \cite{ref5}. So what new opportunities and challenges can it bring to social computing?
\par In this paper, we will give a general overview of the research status of AI and H-AI in social computing, and the challenges facing existing AI. And then we propose a framework of social computing based on H-AI. The main contributions are as follows:
\begin{itemize}
\item Summarizing the future of H-AI in promoting the society towards a more intelligent direction on the basis of the development of AI and human-machine relationship.

\item Presenting the possible challenges of AI in social computing.

\item Proposing a framework of social computing based on H-AI to solve the problems faced by AI in social computing.
\end{itemize}
\par The remainder of this paper is organized as follows. Section 2 summarizes the development from AI to H-AI. Section 3 surveys the research status of AI in social computing. Section 4 presents the possible challenges of AI in social computing. Section 5 propose a framework of social computing based on H-AI. Section 6 concludes this paper.

\section{RESEARCH ON AI IN SOCIAL COMPUTING}

\par The rapid development of AI has helped to tackle many decades-old problems and different research fields, such as intelligent transportation systems, smart healthcare and business intelligence. Social computing is not an exception of the AI revolution, and social computing research also had witnessed similar transformation. In this section we will discuss the recent advances of AI in social computing, and how AI has helped to revolutionize the social computing researches.

\par In the middle of the 1990s, personal computing transformed into social computing. Personal computing is mainly used for individuals who use information technology. Social computing is a research field that is a cross-integration of social behavior and computing systems. It is about how to use computing systems to help people communicate and collaborate, how to use computing technology to study the law of social operation and development trends \cite{ref13}. In recent years, the research and application of social computing have been highly valued in information science and related interdisciplinary fields. Many universities and high-tech companies in the world are actively engaged in research and development related to social computing. Interdisciplinary research institutions such as the Santa Fe Institute, Research and Google and the HP Social Computing Lab, Harvard, Stanford and Cornell begin to describe the complex phenomena in the social system with complexity science, put forward a series of new theories and conduct new researches. They pioneered new research method such as computational social science. Carnegie Mellon University conducted research on public health, bioterrorism  monitoring, and prediction; the University of Southern California studied the multi-agent military virtual rehearsal environment, IBM and Microsoft's Babble, Loops, Wallop etc. carry out the network community research projects. In 2007, a seminar on computational sociology was held at Harvard University. In 2008, the US military held a seminar on social computing, behavioral modeling, and prediction at Arizona State and the University of Arizona conducted an "intelligence and security informatics" study to address national social security issues. On the basis of those researches, Science published a computational sociology article \cite{ref15}, making the intersection of computational science and social sciences ¡ª¡ª social computing is becoming a hot topic of research and application at the forefront of international attention. In 2016, the University of Southern California and the University of Northern California established two new AI research centers to provide solutions to social problems, and built the value of human beings into the design of AI.

\par Since the concept of social computing was introduced in 2000, Chinese scholars have also carried out many studies and achieved remarkable results. In 2005, the Chinese Academy of Sciences established the Innovation Team International Partner Program project and began research on social computing for intelligence security. In 2008, the project "Basic Theory and Prototype System for Information and Security Informatics and Social Computing" was launched to support related theoretical methods and system research and development [10]. Besides, Chinese scholars have conducted in-depth discussions on social computing related issues in social computing conferences and special conferences organized by them. Seminars on "International Social Security and Social Computing" and "Information and Security Informatics Seminar" were held respectively in 2005 and 2006. In 2007, the Xiangshan Science Conference was organized with the theme of "Basic Theory and Application of Social Computing". In 2008, the Academic Salon of the Chinese Association for Science and Technology and the seminar on "Social computing Behavioral Modeling and Forecasting" with the theme of "Social computing-Social Computation?" were organized. In the same year, the first social computing seminar(S0C0 2008) was successfully held, and in 2009, the International Social Computing Conference (ICSC) was convened. After that, the National Social Computing Conference is held every year. In 2012, the "4th National Social Computing Conference" was successfully held to explore the significance of social computing for the development of modern society and to look forward to the future development trend of social computing. In 2015, China also called the "International Intelligence Space Conference" to discuss the most popular topics in the computer field ¡ª¡ª universal and intelligent computing, cloud computing, social computing, etc., to achieve the integration of the four spaces of physical space, cyber space, social space and thinking space. In 2018, the "13th National Conference on Computer Supported Collaborative Work and Social Computing" was held to explore the intelligent collaborative computing research in the physical world, human society and information space and their integration. In August 2019, "the 4th National Conference on Big Data and Computing" was held to explore the application of big data, AI and social computing in social development and governance. In the same month, the School of Social Sciences of Tsinghua University hosted "the 2019 International Conference on Social Computing" to explore how to use computation to understand the explosive growth of social data and the process of human-machine interaction to form more powerful intelligence.

\par As an interdisciplinary field of emerging disciplines, social computing has received increasing attention from information science and related interdisciplinary fields. Current research focuses for AI in social computing include social network analysis, data management, and user interaction research, and group intelligence.

\subsection{Social Network Analysis}

\par Social network is defined as a social structure composed of social individuals on the information network, including three elements: relationship structure, network group, and network information. The relationship structure is the carrier of social network. The network group is the main body of social network, promoting the dissemination of network information. The network information is the starting point and destination of social network, affecting the change of relationship structure, and is the object of social network \cite{ref17}. At present, social network applications are in a period of vigorous development, and the analysis of social network has become a hot issue in the research community.

\par Social network research includes social network structure modeling, personal and group behavior pattern mining, virtual community discovery, information dissemination model, impact analysis, etc. The research results on these issues are very significant. For example, Watts and others verified the "six degrees of separation" and the small world model on a 60000-node mail network \cite{ref18}. McCandless et al. \cite{ref19} proposed a method of information propagation based on Monte Carlo using probability theory. The similarity-based aggregation algorithm proposed by Shen et al. \cite{ref21} is used for the discovery of dynamic virtual communities. The theory of planned behavior proposed by lcek Ajzen \cite{ref23} is widely used in human behavior research.

\par Social network have four characteristics: speed, invasiveness, equality, and self-organization. Because of these characteristics, they also cause many problems to society. For example, users' personal information is in danger of being leaked with involved in social network. This leads to frequent scams. We need to study social network in depth, build a huge functional platform, mine the user behavior of individuals and groups behind the data, study the law of social development, and make better services for society.

\subsection{Data Management}

\par Social computing usually requires a large variety of sensing devices and mobile devices, combined with various social software such as Facebook, MSN, WeChat, etc. to get objective, continuous and dynamic data on human social behavior and interaction. It is the basis for social network analysis to uncover the hidden relationships between people or the communities in which they exist. The characteristics of data collection are:

\begin{enumerate}[(1)]

\item Massive isomerism. From the physical society and the network society, the amount of data of different media types, such as text, image, video, audio, etc. increases exponentially with time. How to establish an effective large-scale, dynamic incremental data integration model and how to establish an efficient retrieval and analysis mechanism are the issues that need to be studied in the field of social computing.

\item Dispersibility. The sources of data are diverse and highly dispersed, and the multi-source nature of massive data poses great challenges for data fusion technologies. And how to integrate multi-source data should be the most fundamental problem in the field of social computing.

\item Uncertainty. A large amount of false, outdated, inconsistent, incomplete data emerges due to the timeless of multi-source data, user input errors, false information deliberately posted by malicious users, etc. But the accuracy, completeness, consistency, credibility, timeliness and other factors of the data directly determine the quality of the data. Data quality is the foundation of all data analysis. How to quickly assess the quality of data and achieve rapid cleaning of data is the most urgent research problem in the field of social computing, and also the basis for conducting social computing research.

\end{enumerate}

\par The above-mentioned characteristics of these data have caused many problems to the study of social computing. However, in recent years, the emergence of big data has brought new opportunities to social computing in the field of data management. Its arrival has had an unprecedented impact on the breadth, depth, and scale of data collection and analysis. Under the conditions of big data, the traditional computation-centered concept gradually shifts to data-centricity and forms data thinking \cite{ref24}.Social computing is a data-intensive scientific research paradigm. Data mining based on content information and social computing based on structural information is a research hotspot in the field of big data mining and social computing \cite{ref26}. Many international universities have done a lot of research on data. For example, James A. Evans explains how to use data to understand society and analyze how complex, dynamic, adaptive social systems and human-machine interaction from more powerful intelligence. Data management should always be a hot issue in the field of social computing research, in which it continues to make theoretical and technological breakthroughs.

%\IEEEpubidadjcol

\subsection{User interaction research}

\par From the user level, AI in social computing research is abut how to promote user-user interaction, and social impact analysis of users shown through user interaction. Whether it is Web 2.0 or a virtual social network system, it emphasized the interaction between users, and realizes the interconnection of people. Traditional human-machine interaction emphasizes the optimization of software application and interface through program personnel to increase the friendliness of the system. The interaction between people is more about how to realize the interconnection of people and realize information exchange and knowledge sharing. Now the development of AI has further promoted the research of social computing in user interaction. For example, the development of voice interaction technology has changed the mode of human interaction from "Finger-first" to "Voice-first", and the naturalness of voice interaction has been further enhanced, and the experience of human natural dialogue has become increasingly popular. Multi-channel fusion interaction, intelligent person-to-person interaction and other technologies make it possible to give machine context awareness and self-awareness, which makes it possible to build a machine-active interaction model for human and further enhance the experience level of human-machine interaction \cite{ref54}. In the future, emotional computing and cognitive computing in the field of social computing will be the key technology to promote user interaction research.

\subsection{Group wisdom}

\par Group wisdom is machine-assisted and solves problem by means of group collaboration. A typical case of group intelligence is the reCAPTCHA system. Luis Von Ahn presented the "RecAPTCHA: Human-Based Character Recognition vis Web Security Measures" in 2008 using a network authentication code to identity characters by the power of the user \cite{ref58}. The algorithms of group intelligence are mostly based on AI technology, such as particle swarm optimization algorithm, artificial ant colony algorithm, fireworks algorithm, etc. It is currently an issue of social computing. Take the crowd as the research object and theoretical explanation of the human cluster behavior will further promote social research.

\subsection{Social media marketing}

\par With the growth social media content to the scale of big data, it is almost impossible for conventional methods of social media analysis to extract useful knowledge from the user generated big data. Here comes the role of AI and machines learning techniques that can be trained to leverage user preferences, interests, emotions and behaviors to offer him the best personalized experience. Analyzing social media content using AI enabled algorithms can help understand the user's behavior and even intention. It can even predict where the user is going, what he have written in emails, where he have been, what he have asked his voice assistants, what groups he belongs to, what stores he shops at, and more. AI extracted knowledge becomes a great source for social media marketing. The snowball of applying AI in social media marketing has exponentially increased, and it is expected to continue to grow to reach a market value of USD 2.1 billion by 2023 \cite{59}.

\subsection{Personality computing and affective computing}

\par With the proliferation of online social networks, more and more people share their daily stories and memories on social platforms, and the capacity of human analysis lags far behind the amount of user-generated content. Therefore, the need for artificial intelligence methods that can analyze huge amounts of data has led to the emergence of new research areas, such as personality computing and affective computing. Personality computing is a research area that concerns the application of artificial intelligence and personality psychology by computational techniques from different sources, including natural language processing, multimedia analysis and social networks mining. Affective computing (also called artificial intelligence, or simply emotion AI) is the research area that studies the development of systems that can process, recognize, interpret and simulate human effects such as emotions, or mood and preferences. Affective computing is an interdisciplinary research field incorporating computer science, cognitive science and psychology.

\par In-depth research on social computing has been conducted. Table I summarizes the results of social computing research in recent years.

\begin{table}
    \caption{Hot topics of social computing and representative solutions}

    \begin{tabular}{|p{2cm}|p{6cm}|}
    \hline
    Hot topics & Solution \\
    \hline
    Social network structure modeling & A multi-graph intersection model based on node uniqueness identification \cite{ref32}.\\&super-graph modeling method \cite{ref33}. \\
    \hline
    Individual and group behavior pattern mining & A measure of attribute uniqueness based on association analysis \cite{ref34}.\\&Established individual migration and group migration patterns for seven social networks.\cite{ref35}\\&A location-based social network user behavior analysis method. \cite{ref36} \\
    \hline
    Virtual community mining & Divide the network community algorithm by finding the network with the largest modularity in the community. \cite{ref37} \\&Study the discovery of network-level communities. \cite{ref38}\\
    \hline
    Impact analysis & Use probability theory to propose Monte Carlo-based information propagation calculation method. \cite{ref39} \\&An influence evaluation function method based on entropy for network information propagation.\cite{ref40} \\
    \hline
    Information dissemination model & SIR model based on small world network. \cite{ref41} \\&BRPT algorithm. \cite{ref42} \\
    \hline
    Use interaction research & Speech recognition technology based on vocal tract model and speech recognition method, template matching method and neural network method. \cite{ref47}, \cite{ref48} \\ &Multi-channel interaction technology. \cite{ref49}, \cite{ref52}.\\&Agent-based human-computer emotion interaction technology. \cite{ref53}\\
    \hline
    Data Management & Method for time clustering.\cite{ref43} \\& Study label learning and ranking in social labeling systems \cite{ref44},\cite{ref45}, information extraction and classification. \cite{ref46}\\
    \hline
    \end{tabular}

\end{table}

\section{THE INTELLIGENCE DEVELOPMENT \\ FROM AI TO H-AI}
\par AI has gone through several development periods since it was proposed by McKay in 1956. The development of AI is shown in Figure 1.

\begin{figure*}[hbt]
\centering
\includegraphics[width=15cm]{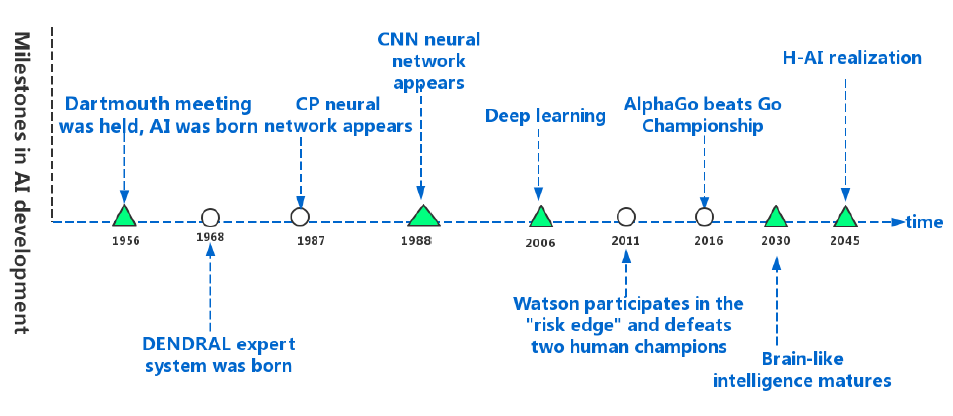}
\caption{From AI to H-AI Timeliness \cite{ref6}- \cite{ref8}}
\end{figure*}

\par In generally, we will find traces of AI in all aspects of our daily lives. Its application has become very extensive, including computers, voice search, face recognition, mobile phones, autonomous vehicles, etc. Although widely used, AI can be divided into three levels according to the degree of intelligence. The classification is shown in Table II.

\begin{table}[h]
    \caption{The classification of AI}
    \begin{tabular}{|c|p{2.2cm}|p{2cm}|p{1.3cm}|}
    \hline
    Level & Description & Instance & Scope of application \\
    \hline
    Elementary AI & Low level intelligence & Automatic interception of harassing calls, automatic filtering of email addresses, Watson, AlphaGo & Large\\
    \hline
    Advanced AI & Intelligent machines that can solve problems, have excellent learning ability and improve themselves through learning. & Sweeping robot, autonomous vehicle, unmanned aerial vehicle. & Medium\\
    \hline
    Super AI & Surpassing the level of human intelligence & \center{---} & None\\
    \hline
    \end{tabular}
\end{table}

\par With the development of AI, the relationship between AI and human become close. Therefore, it prompts us to think about man-machine relationship. The man-machine relationship will experience three levels: man-machine interaction, man-machine cooperation, and man-machine integration. Man-machine interaction refers to the process of information exchange between man and machine using a certain dialogue to complete the specific task in a certain interaction manner. Traditional interaction methods include keyboards, mice, touchpads, etc. Nowadays, the rise of virtual reality augmented reality and somatosensory technology has further expanded the relationship between people and machines. Man-machine collaboration refers that machines assist human to deal with a large number of complex and highly accurate jobs. Man and machines develop an interdependent relationship. Man-machine fusion refers to the integration of human intelligence and AI. This relationship is not a simple "human + machine", but a new type of intelligent system. Human beings and machines become a symbiotic coexistence relationship. At present, the man-machine should be the relationship of human-machine cooperation, and the transition from man-machine cooperation to man-machine fusion has not yet been realized. However, the dual intelligence generated by man-machine fusion can become the top priority of AI in its future development. AI becomes the natural extension and expansion of human intelligence, and complex problems can be solved more efficiently. It has profound scientific significance and great industrialization prospects \cite{ref5}.

\par Because of the interaction between human and machine, it generates a new type of intelligence called Human-artificial intelligence (H-AI). It is a new intelligent generation that combines physical and biological factors which differentiates  from human intelligence and artificial intelligence \cite{ref11}. Human intelligence is more efficient than AI in terms of perception, reasoning, and learning. However, the search, calculation and storage capabilities of machines far exceed human intelligence. The future H-AI will combine the advantages of human beings and machines to realize the deep integration of human cognitive attributes and machine computing attributes, forming a new enhanced intelligence and flexibly solve various contradictions and paradoxes in the issue of H-AI. H-AI is a two-way closed-loop system, as shown in Figure 2. It can realize the "perception-decision-behavior-feedback" closed-loop workflow, in which people can accept machine information, machines can also read human signals, and they can promote each other to shape a more efficient form of human-machine interaction.

\begin{figure}[!hbt]
\includegraphics[height=5cm,width=8cm]{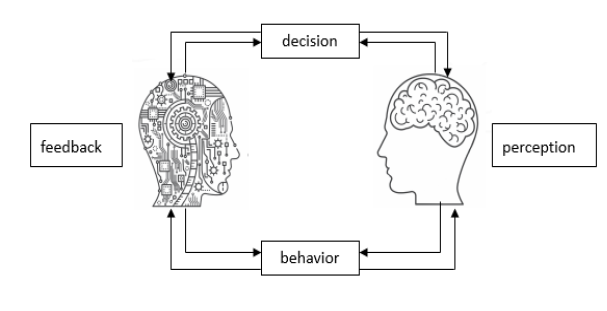}
\caption{H-AI bidirectional closed-loop system}
\end{figure}

\par The concept of H-AI combines the concept of human intelligence and machine intelligence, and it provides a new method for dealing with incomplete or unstructured data through human-machine interaction. Social data is currently accumulated from blogs, wikis, multi-player games, and other social media media tools in an unprecedent way, and it is important to mine hidden information and analyze the social implications behind social media. And H-AI can help improve the ability to mine and analyze hidden information, including behavior recognition, intent prediction, etc. Exploring H-AI for social computing can also improve decision making and help with social perception, social modeling, resource scheduling, behavior analysis, using computing systems to help people communicate, and study the laws and trends of social operations.

\section{THE CHALLENGES OF AI IN SOCIAL COMPUTING}

\par The rapid development of AI has brought many challenges to the field of social computing while achieving a series of remarkable results. The following is a detailed analysis of the problems existing in social computing.

\subsection{Analysis problem pattern}

\par The core technology of AI is machine learning. It is to let computers user a specific algorithm such as decision tree algorithm, artificial neural network, Bayesian algorithm, etc. to train the model behind these data onto a given data set. When new data emerge, people can use these models to make predictions and analysis. Although some social behavior of human beings is highly modeled, for example, given a particular resource, individuals typically make decisions that maximize their benefits. At this point, the machine analyzes this modeled behavior, based on a large number of existing experience, to provide individuals with optimal decisions, which are often overlooked by human when making their own decisions. From this perspective, AI can indeed help human make more reasonable judgements. However, if this method is used to understand and study society, it will not work.

\par Although AI can perform algorithm-based prediction and analysis on modeled behavior, once people are faced with non-model problems, such as the study of friendly relations between people in the field of social computing, people's preferences will change with time, using AI to solve  non-patterned and dynamic problems in these societies. These problems have not yet achieved the desired results.

\subsection{Lack of systematic abstract thinking}

\par It is now possible to use natural language processing techniques to analyze the author's emotions expressed in a piece of text, but it is still relatively mechanical, and the specific words in it have the computational characteristic. So it is very preliminary for people to get the information about how to use this technology. In other words, it is currently difficult for people to expect that AI can read the potential meaning between lines. However, in social computing research, it is precisely a paragraph of words that can express people's real ideas. If patterns based on word frequency or other superficial words are followed, it is difficult to distinguish between people's true meaning and irony, because in general, a Chinese text or Chinese character string may have multiple meanings. Conversely, an identical or similar meaning can also be represented by multiple Chinese texts or multiple Chinese character strings, which are also the main difficulties and obstacles in natural language understanding.

\par At present, AI in social computing is lacking in abstract thinking, and its dealing with specific problems is a separate analysis for lack of holistic and systematic thinking. In the field of social computing, what interests researchers is the connection between people, and the society of independent existence that is generated by this connection and simple summation beyond individuals.

\subsection{Unfriendly service}

\par The rise of intelligent services such as intelligent customer service and chat robots can help people solve some conventional problems, but they still have many shortcomings, and criticisms are constantly emerging. They are described as dumb, frustrating, useless, totally time-consuming chat machine, characterized as machines that are unfriendly lacking in humanity. The future of chat bots should be seamless, that is, users can not tell whether it is a robot or a human, which is the ultimate goal of AI services.

\par At present, people and machines are in a low-level collaboration and cannot provide high-quality services. The unfriendly service has become a problem worthy of attention in the field of social computing applications. How to optimize services? How do human and machines achieve high-level collaboration? This will be the focus of the future research.

\subsection{Personal privacy issues}

\begin{figure*}[htp]
  \centering
  \includegraphics[width=15cm]{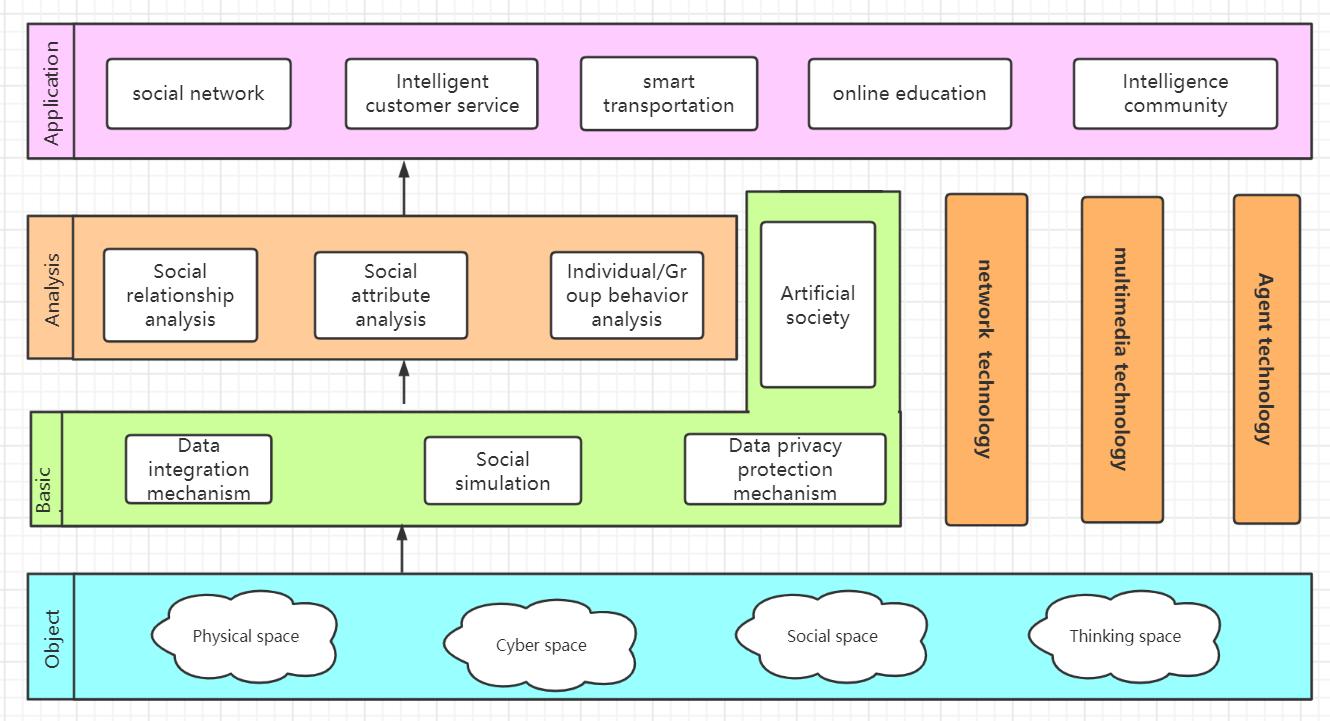}

  \caption{The framework of social computing based on H-AI}
\end{figure*}
\par With the rapid development and popularity of emerging information technologies such as the Internet, cloud computing, the Internet of Things and social networks, data is growing at an explosive rate and deeply integrated into all aspects of human society. The data on social computing is closely related to people. For social computing research, open-source data is needed to better explore human behavior, group relationship, and social issues. This also avoids the growing problems of security attack and privacy breaches arising from data open-source.

\par The development of social media and social networks has left personal privacy data, such as personal consumption records, economic status, medical information, etc., more easily acquired and spread around the world at an unexpected rate. From this point of view, the open-source and confidentiality of data forms a pair of seemingly irreconcilable contradictions. Open-source makes data accessible to people in an open way, and realizes knowledge sharing and scientific research, and confidentiality calls for the protection of individuals and organizations from publicity. Sensitive information, social computing depends on both data open source and data protection \cite{ref23}.

\par The issue of privacy protection faced by social computing has always been a matter of great concern to the academic community. If a breakthrough is made on this issue, the research on social computing is more mature, and the research on mining human behavior and social relations is deeper.

\par Society is a complex and dynamic system. We cannot hinder the development of science and technology innovation because of the challenges from new social phenomena. However, we can achieve the regulation of human-machine hybrid behavior through the combination of social science and computer science. This should be the road of scientific development that we seriously consider.

\section{Framework Proposed based on H-AI}
\par Many problems faced by AI in the field of social computing have been introduced in the previous section. Then, can H-AI, combined with the advantages of human intelligence and machine intelligence, solve problems in social computing? The H-AI for social computing framework proposed to the problems raised above in this paper, as shown in Figure 3. This framework consists of four parts: the object layer, the base layer, the analysis layer, and the application layer.

\begin{itemize}
  \item Object layer
  \par The objects of social computing mainly include the traditional society and the virtual society. Wang et al. \cite{ref50} proposed the cyber-physical-social systems (CPSS). The system integrates people and society into physical and cyberspace, making it possible to interact with virtual reality, closed loop feedback, and parallel execution. At the same time, the advantages of human intelligence and machine intelligence can be brought to solve complex problems in society in this system. In addition, the document \cite{ref25}, \cite{ref56} proposed a social computing method based on ACP, and analyzed and discussed the complex problems of CPSS. In fact, in a narrow sense, social computing is a massive information resource generated by human beings through physical society and the network society. Through data mining, a corresponding model is established to predict what kind of behavior individuals or groups will have and how to change their behavior.
  \item Base layer
  \par The degree of user participation on the internet has greatly increased while leaving a large number of social footprints. Due to the multi-source nature of the data, it is necessary to integrate a large amount of data from each data source, so the data integration mechanism is introduced in the base layer.
  \par As AI develops, social simulation technology has become increasingly mature, and significant progress has been made in modeling ideas and technology empowerment. The theory of complex adaptive (CAS) proposes that the emergence of the subject-based modeling (ABM) method, distributed AI, genetic algorithm (GA), cellular automata (CA) and the birth of neural networks makes the social simulation more realistic in imitating the real phenomena in social life. But can social simulation simulate social systems? What is the foundation of its plausibility? If these problems are not well resolved, they will constrain the application of the social sciences. If based on social simulation, combined with H-AI, the machine that allows the owner¡¯s ability to be the object of social simulation, let it interact on the computer platform, then the plausibility of social simulation will be greatly improved, and various behavior of the research object will be more convincing. The private system is also included in the base layer. Personal privacy is one of the main problems in social computing research. In order to solve the data privacy problem of social computing, Yuan Yong et. al. \cite{ref22} proposed that it should integrated with the blockchain to realize the full mining and sharing of data under the premise of protecting data privacy. Blockchain digital encryption technology can encrypt data to prevent data from being abused by anyone other than the person before it is authorized. Moreover, the blockchain can trace the traceability of information and record data, and protect personal data in real-time.
  \item Analysis layer
  \par This layer includes social interpersonal relationship mining, social attribute analysis, and social group behavior analysis. Typical method studies have been given in Table I. The computer only mines the objective data obtained and analyzes the problem model. The society is a complex system and has various kinds of non-pattern problems. The problem of pattern is transformed into a problem that the machine can handle. The fusion of human intelligence and machine intelligence will analyze the social operating laws and social phenomenon that have not been discovered in society.
  \item Application layer
  \par It applies the theory and method of social computing to the actual society. Combined with AI, social computing has its typical applications including social networks, intelligent customer service, intelligent transportation, online education, and intelligent society. In spite of the extensive application, social computing has many potential problems. Supposing that the machine can think like humans, in the application layer of social computing, the advantages of human intelligence are exerted, the ability of human and machines is qualitatively leaped. A smart society is based on a massive social sensor network, supported by high-performance distributed computing, combined with knowledge automation technology, using the parallel system of virtual and real interaction as a means to promote comprehensive perception, modeling, analysis, decision-making and feedback execution of society, and in turn, it realizes the social management of closed-loop feedback, which integrates emotional knowledge, parallel management and mobile command and control, and comprehensively enhances its function and services \cite{ref25}. H-AI plays an important role in all layers of the social computing framework. The advantage of machine intelligence and human intelligence integration in social computing is unmatched by AI. The development of dual intelligence will promote the explosion of social computing development.
\end{itemize}

\section{Conclusion}

\par Now, the third wave of AI is ushered in. With the development of science and technology, the future of AI will transit to the intelligence stage of human-machine integration. H-AI is likely to be an advanced stage of AI, and may even be its final stage. The integration of people and machines will complement each other and promote the rapid development of social computing and solve bottlenecks in scoial computing research. And if a "hand in hand" development of AI and social computing is achieved, we can not only describe the process and pattern of information interaction between people more clearly, but also explain phenomena and its mechanism more reasonably, and even accurately predict the daily behavior of people and the development trend of social systems, to jointly design and build a better intelligent society.

% or


\section*{Acknowledgment}

This work was supported by the National Science Foundation of China under Grant 61872038, 61811530335, and in part by the UK Royal Society-Newton Mobility Grant (No.IECnNSFCn170067) and the Fundamental Research Funds for the Central Universities under Grant FRFBD-18-016A.

\ifCLASSOPTIONcaptionsoff
  \newpage
\fi



%\bibliographystyle{IEEEtran}

\begin{IEEEbiographynophoto}{Wenxi Wang}
received her B.E. degree from Ludong University in 2019. She is currently pursuing the M.S. degree in the School of Computer and Communication Engineering, University of Science and Technology Beijing. Her current research interests include Social computing and Artificial Intelligence.
\end{IEEEbiographynophoto}

\begin{IEEEbiographynophoto}{Huansheng Ning}
received his B.S. degree from Anhui University in 1996 and his Ph.D. degree from Beihang University in 2001. He is currently a Professor and Vice Dean with the School of Computer and Communication Engineering, University of Science and Technology Beijing, China, and the founder and principal at Cybermatics and Cyberspace International Science and Technology Cooperation Base. He has authored 6 books and over 150 papers in journals and at international conferences/workshops. He has been the Associate Editor of IEEE Systems Journal, the associate editor (2014-2018) and the Steering Committee Member of IEEE Internet of Things Journal (2018-), Chairman (2012) and Executive Chairman (2013) of the program committee at the IEEE international Internet of Things conference, and the Co-Executive Chairman of the 2013 International cyber technology conference and the 2015 Smart World Congress. His awards include the IEEE Computer Society Meritorious Service Award and the IEEE Computer Society Golden Core Member Award. His current research interests include Internet of Things, Cyber Physical Social Systems, electromagnetic sensing and computing. In 2018, he was elected as IET Fellow
\end{IEEEbiographynophoto}

\begin{IEEEbiographynophoto}{Feifei Shi}
received her B.S. degree from China University of Petroleum in 2016 and her M.S. degree from University of Science and Technology Beijing. She is currently a Ph.D. student in the School of Computer and Communication Engineering, University of Science and Technology Beijing. Her current research interests include Internet of Things and Artificial Intelligence.
\end{IEEEbiographynophoto}

\begin{IEEEbiographynophoto}{Sahraoui Dhelim}
received his B.S. in Computer Science from the University of Djelfa, Algeria, in 2012 and his Master degree in Networking and Distributed Systems from the University of Laghouat, Algeria, in 2014. Since 2015 he has been pursuing his PhD at the University of Science and Technology Beijing, Beijing, China. He is an active reviewer in many journals, including IEEE Transactions on
Computational Social Systems, IEEE Transactions on Intelligent Transportation Systems and IEEE Transaction on Vehicular Technology. His current research interests include Social Computing, Personality Computing, User Modeling, Interest Mining, Recommendation Systems and Intelligent Transportation Systems.
\end{IEEEbiographynophoto}

\begin{IEEEbiographynophoto}{Weishan Zhang}
received the Ph.D. degree from Northwestern Polytechnical University, Xi¡¯an, China, in 2001. He is currently a Full Professor with the Department of Software Engineering, China University of Petroleum, Qingdao, China. He has authored or coauthored
over 100 papers. His current research interests include big data platforms, pervasive cloud computing, and service-oriented computing. His current H-index according to Google Scholar is 15.
\end{IEEEbiographynophoto}

\begin{IEEEbiographynophoto}{Liming Chen}
is a Professor and Research Director in the School of Computing at Ulster University, UK. His research interests include data analytics, pervasive computing, artificial intelligence, user-centred intelligent systems and their applications in digital healthcare. Liming received a Ph.D. in Artificial Intelligence from De Montfort University, UK. He is an IET Fellow, a Senior Member of the IEEE. He has authored 6 books and over 230 peer-reviewed papers.
\end{IEEEbiographynophoto}
\end{CJK*}
\end{document}